\documentclass[letterpaper, 10pt, conference]{ieeeconf}  
\IEEEoverridecommandlockouts                              

\usepackage{booktabs}
\usepackage{graphicx}
\usepackage{array}
\usepackage{cite}
\usepackage{graphicx}
\usepackage{amsmath}
\usepackage{amssymb}
\usepackage{algorithm}
\usepackage[noend]{algorithmic}
\usepackage{flushend}
\usepackage[T1]{fontenc}
\usepackage{aecompl}
\usepackage{xurl}
\usepackage{hyperref}
\usepackage{ifthen}
\usepackage{caption}

\newcommand{\Symbol}[1]{\ensuremath{\mathcal{#1}}}

\newcommand{\Var}[1]{\ensuremath{{{\mathrm{#1}}}}}

\newcommand{\Goal}{\StateSpace_\textit{goal}}

\newcommand{\StateSpace}{\mathcal{S}}
\newcommand{\ActionSpace}{\Symbol{U}}
\newcommand{\MotionEqs}{\ensuremath{f}}
\newcommand{\AngularVelocity}{\ensuremath{w}}
\newcommand{\LinearVelocity}{\ensuremath{v}}

\newcommand{\CostFunction}{\Gamma}

\usepackage{xcolor}
\definecolor{green}{RGB}{11,155,13}

\title{\bf Decremental Dynamics Planning for Robot Navigation}

\author{Yuanjie Lu$^1$, Tong Xu$^1$, Linji Wang$^1$, Nick Hawes$^2$, and Xuesu Xiao$^1$
\thanks{$^1$Department of Computer Science, George Mason University, Fairfax, VA 22030, USA.}%
\thanks{$^2$Department of Engineering Science, University of Oxford, Wellington Square, UK}
}

\begin{document}

\maketitle

\begin{abstract} 
Most, if not all, robot navigation systems employ a decomposed planning framework that includes global and local planning. To trade-off onboard computation and plan quality, current systems have to limit all robot dynamics considerations only within the local planner, while leveraging an extremely simplified robot representation (e.g., a point-mass holonomic model without dynamics) in the global level. 
However, such an artificial decomposition based on either full or zero consideration of robot dynamics can lead to gaps between the two levels, e.g., a global path based on a holonomic point-mass model may not be realizable by a non-holonomic robot, especially in highly constrained obstacle environments.   
Motivated by such a limitation, we propose a novel paradigm, Decremental Dynamics Planning (DDP)\footnote{\url{https://github.com/yuanjielu-64/barn_challenge_lu.git}}, that integrates dynamic constraints into the entire planning process, with a focus on high-fidelity dynamics modeling at the beginning and a gradual fidelity reduction as the planning progresses.  
To validate the effectiveness of this paradigm, we augment three different planners with DDP and show overall improved planning performance. We also develop a new DDP-based navigation system, which achieves second place in both the simulation phase and real-world phase of the 2025 BARN Challenge\footnote{\url{https://cs.gmu.edu/~xiao/Research/BARN_Challenge/BARN_Challenge25.html\#leaderboard}}. Both simulated and physical experiments validate DDP's hypothesized benefits.

\end{abstract}


\section{Introduction}
\label{sec:Intro}

Navigation is a fundamental capability for autonomous mobile robots, enabling them to effectively traverse complex environments without collisions. As the demand for robotic systems grows across various domains, such as industrial automation, search and rescue, and autonomous delivery, the need for efficient and robust navigation strategies becomes increasingly important.

\begin{figure}[!t]
\centering
\includegraphics[width=0.95\columnwidth]{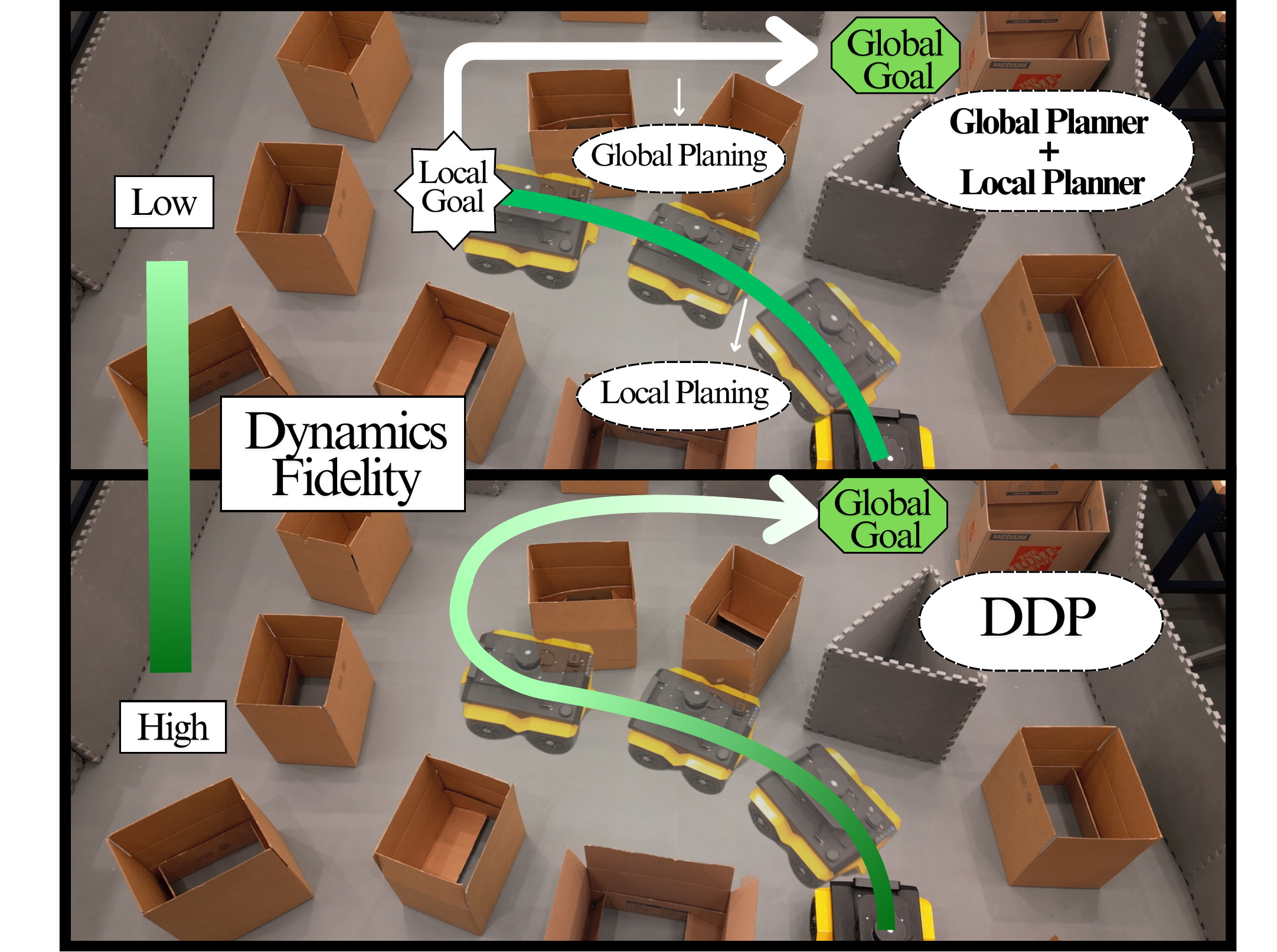}
\caption{Contrasting the traditional global and local planning paradigm (top), where either full (green) or zero (white) robot dynamics is considered, DDP starts with high fidelity dynamics in the early part of trajectory rollout and gradually decreases dynamics fidelity for computation efficiency (bottom).}
\label{fig:DDP}
\end{figure}

Traditionally, most robot navigation systems adopt a hierarchical planning framework, decomposing the planning process into global and local planning. Global planning aims to quickly plan a high-level path from the robot's current position to the desired goal, while local planning focuses on tracking the global plan while providing real-time, fine-grained obstacle avoidance and path execution in the vicinity of the robot~\cite{pandey2017mobile}. To balance limited onboard computation, existing systems typically only consider the robot's dynamics in the local planner, while using a simplified robot representation, such as a point-mass holonomic model, in the global planner~\cite{xiao2022motion}. 

Although this decomposed architecture is widely used, the artificial separation of global and local planning, based on either full or no consideration of robot dynamics, can lead to significant differences between the two levels. For instance, a global path generated for a holonomic, point-mass robot may guide it through a narrow corridor with lots of turns. However, the real, physical robot, with its specific kinodynamic constraints, might struggle to navigate through this passage. 
More generally, since the global planner fails to incorporate robot dynamics, it often produces non-smooth paths with sharp turns, routes passing through challenging areas, or plans that are infeasible.

To overcome this limitation of traditional navigation planning, we propose Decremental Dynamics Planning (DDP), a novel paradigm that seamlessly integrates dynamic constraints into the entire planning process, eliminating the need for a decomposed framework. Specifically, DDP starts with high-fidelity dynamics modeling in the early trajectory rollout stages, capturing essential dynamic properties of the robot, e.g., velocity, acceleration, and turning radius constraints, and ensuring that the robot can precisely navigate complex environments and avoid highly constrained obstacles. As the trajectory rollout progresses, the fidelity of dynamic modeling gradually decreases by simplifying the model to improve computational efficiency while ensuring that dynamic feasibility is not significantly compromised (Fig.~\ref{fig:DDP}). 

To validate the feasibility and effectiveness of this paradigm, we use DDP to enhance three different planners, namely DWA~\cite{fox1997dynamic}, MPPI~\cite{williams2016aggressive}, and Log-MPPI~\cite{mohamed2022autonomous}. Experimental results demonstrate that, compared to the baseline methods, DDP significantly improves planning success rates and efficiency, in both simulated
and physical experiments. In addition, we develop a new standalone DDP-based navigation system, which achieves second place in both the simulation and real-world phases of the 2025 BARN Challenge.










\section{Related Work}
\label{sec:RelatedWork}

\subsection{Autonomous Robot Navigation}
Autonomous robot navigation involves several research areas: perception, state estimation, motion planning, and motion control. Most navigation systems follow a hierarchical planning paradigm: global planning and local planning. 

Global planning typically employs grid-based methods~\cite{elfes1989using, marder2010office, lu2014layered} or sampling-based methods~\cite{kavraki1996probabilistic, lavalle1998rapidly} to construct tractable environment representations. To further simplify computations considering the long-horizon, expensive global planning problem being solved onboard a mobile robot, the robot is often modeled as a holonomic point mass without considering dynamics~\cite{khanal2023learning}. Based on this assumption, common search algorithms, e.g., A$^*$~\cite{hart1968formal}, Dijkstra's algorithm~\cite{dijkstra2022note}, and their variants~\cite{pivtoraiko2011kinodynamic, likhachev2009planning, honig2022db} compute the shortest path. However, this simplification overlooks robots' dynamic constraints, potentially resulting in infeasible or suboptimal paths.

Given a coarse global path, local planning takes a local goal or a small local portion of the global path as input and refines the trajectory based on current sensory data, generating real-time motion control commands that ensure obstacle avoidance. By forward-simulating the robot's dynamics, methods like DWA~\cite{fox1997dynamic,  xiao2022appl} and MPPI~\cite{mohamed2022autonomous, williams2017model} generate real-time motion commands by sampling feasible low-level actions to ensure obstacle avoidance. 
Although local planning considers dynamic constraints, the high computation demand of unfolding robot dynamics limits its planning to short horizons. Therefore, coordination between the artificially separated global and local planning becomes a challenge. The prevalent binary treatment of zero or full dynamics in global and local planning respectively often leads to significant gaps between the two levels and makes navigating complex areas difficult~\cite{xiao2023autonomous, xiao2024autonomous, wang2024grounded}, e.g., a global path may erroneously lead the robot into a narrow, twisting passage which the local planner cannot maneuver through. 

\subsection{Robot Dynamics}
Robot dynamics studies how forces and torques influence a robot's motion, considering factors such as mass, inertia, friction, and external forces. Unlike kinematics, which focuses solely on describing motion, dynamics ensures that the planned movements are physically feasible and realistic. However, generating optimal trajectories under dynamic constraints presents significant computational challenges. 


Commonly-used ground robot dynamics models include bicycle, snake-like, Ackermann-steering, and differential-drive models~\cite{lu2025multi, das2024motion, lu2023leveraging, bui2022improving}. More sophisticated dynamics models that consider environment features have also been developed for high-speed navigation~\cite{pokhrel2024cahsor, xiao2021learning, karnan2022vi} and mobility on vertically challenging terrain~\cite{datar2024toward, xu2024reinforcement, datar2024terrain, datar2023learning, cai2025pietra} in off-road operations. 
These approaches combine classical and machine learning methods~\cite{xiao2022motion} to enable more precise trajectory prediction. 
Although high-fidelity dynamics models can precisely predict future robot states so planners can confidently generate actions, they also demand substantial computational resources, posing challenges for real-time onboard applications.

DDP aims to address the binary treatment of robot dynamics in classical decomposed navigation systems and to efficiently integrate dynamic constraints into the entire planning process. DDP maintains high-fidelity dynamic modeling in the early stage of trajectory rollout, strictly adhering to the robot's dynamic characteristics to ensure precise navigation through complex environments and effective obstacle avoidance. As the trajectory progresses, the fidelity of dynamic modeling gradually decreases, eventually reaching a point where path planning is performed entirely based on a holonomic point-mass model without considering dynamics.

\section{Decremental Dynamics Planning}
\label{sec:Problem}
We first present a motion planning problem with robot dynamics, which is usually intractable to solve for long-horizon navigation tasks in real time. We then illustrate current practice to solve this, i.e., decomposing the problem into global and local planning. Our presentation reveals the limitations of such a decomposition. DDP aims to mitigate these. 

\subsection{Original Motion Planning Problem Formulation}
We define the robot state space $\StateSpace = \StateSpace_\textit{free} \cup \StateSpace_\textit{obs}$, where $\StateSpace_\textit{free}$ and $\StateSpace_\textit{obs}$ denote free spaces and obstacle regions. The robot control space is defined as $\ActionSpace$ and the robot dynamics model $\MotionEqs$ is expressed as forward dynamics in the form
\begin{equation}
s_{t+1} = f(s_t, u_t; \theta), s_t \in \StateSpace, u_t \in \ActionSpace,
\label{eqn::original}
\end{equation}
where $\theta$ represents the parameters of the robot's dynamics model, e.g., mass, inertia, wheelbase, axle length, integration interval, or, in the case of a learned dynamics model, parameters of the learned model. Some of these parameters in $\theta$ determine how precise $f$ is in representing the physical robot. In most cases, $\theta$ balances a tradeoff between the fidelity and query speed of $f$, which significantly affects the efficiency of a navigation planner onboard a computation-limited mobile robot platform. 

Using ground robots as an example, the state $s_t = (x, y, \psi)$ is defined by the translations along the $\mathbf{x}$ and $\mathbf{y}$ axis ($x$ and $y$) and the rotation along the $\mathbf{z}$ = $\mathbf{x} \times \mathbf{y}$ axis (yaw $\psi$) of a fixed global coordinate system at time step $t$. For some robot models (e.g., Ackermann-steering), the state may also include additional dimensions such as steering angle $\delta$ and linear velocity $v$. The control $u_t$ is defined by the robot's physical characteristics. If the robot is a differential-drive robot, the control $u_t$ can be defined as the linear velocity $\LinearVelocity$ and angular velocity $ \AngularVelocity$. In contrast, if the robot is based on Ackermann-steering, the control $u_t$ can be defined as the acceleration $\mu_\Var{acc}$ and steering rate $\mu_\Var{steer}$. 

Finally, given an initial state $s_0$ and a goal region $\Goal \subset \StateSpace$, the motion planning problem is to find a control function $u: \{t\}^{T-1}_{t=0} \to \ActionSpace$ that generates a collision-free and dynamically-feasible trajectory $s_t \in \StateSpace_\textit{free}, \forall t \in \{t\}^{T}_{t=0}$ from the initial state $s_0$ to the goal region $\Goal$ and minimizes a given cost function $\CostFunction(s)$, which maps from a state trajectory $s: \{t\}^{T}_{t=0} \to \StateSpace$ to a positive real number. The minimum-cost state trajectory according to the cost function $\CostFunction$ over time step $T$ is the optimal solution to the original motion planning problem. However, given that performant navigation requires solving this optimisation at 20Hz or higher, it is impractical to find an optimal trajectory which fully satisfies the robot dynamic constraints in all time steps. Therefore, approximate solutions are necessary to ensure real-time performance. This is often done by decomposing the problem and simplifying robot dynamics when appropriate. 

\subsection{Conventional Decomposition of Global \& Local Planning}
Facing the intractable nature of solving the above problem on a mobile robot in real time, conventional navigation systems typically decompose the original problem into global and local planning, which consider robot dynamics differently. We represent this decomposition as a choice between parameters $\theta$ for local planning (including dynamics), and $\varnothing$ for global planning (simplified model): 
\begin{equation}
s_{t+1} = 
\begin{cases}
f(s_t, u_t; \varnothing),  & \text{if  } t > \tau \text{ or  } ||s_t-s_0|| > d \text{ (global)},  \\
f(s_t, u_t; \theta), & \text{if  } t \le \tau \text{ or  } ||s_t-s_0|| \leq d \text{ (local)},
\end{cases}
\label{eqn::decomposed}
\end{equation}
where the decomposition happens either at $\tau$, a time step threshold, or $d$, a distance threshold. Both time step or distance threshold define the computational window within which the robot's dynamic model is considered with static parameters $\theta$. In general, $\tau$ or $d$ is manually determined based on the deployment environment and platform. 

Most conventional navigation systems perform global planning first, then use its output as input for local planning, which, e.g., follows it closely using the dynamics model. 
However, this artificial separation of global and local planning leads to dynamically infeasible paths right beyond $\tau$ or $d$, which may cause problem for local planning and reduce overall navigation performance. For example, in highly constrained environments, the global plan without considering dynamics may lead the robot into a narrow and twisty passage, which is impassable when adhering to dynamic constraints. DDP is designed to mitigate the challenges caused by such artificial problem decomposition and binary dynamics consideration by efficiently infusing dynamics into the entire planning process. 

\subsection{Generalization by Decremental Dynamics Planning}
In order to more effectively approximate the full dynamics in the original navigation problem (Eqn.~\eqref{eqn::original}), DDP generalizes the decomposed dynamics model (zero or full dynamics, Eqn.~\eqref{eqn::decomposed}) as:
\begin{equation}
s_{t+1} = f(s_t, u_t;\theta_t),
\label{eqn::ddp}
\end{equation}
where $\theta$, instead of being a constant set of parameters (as in Eqns.~\eqref{eqn::original} and \eqref{eqn::decomposed}), becomes a function $\theta: \{t\}^{T-1}_{t=0} \to \Theta$ where $\Theta$ is the space of dynamics parameters for $f$.   
Notice that Eqn.~\ref{eqn::decomposed} is a special case of Eqn.~\ref{eqn::ddp}, where only two parameter sets ($\theta$ and $\varnothing$) are chosen based on a manually chosen threshold ($\tau$ or $d$). 
To efficiently utilize limited onboard computation, higher fidelity dynamics models should be utilized at the beginning of the trajectory rollout to ensure precision. 
As trajectory rollout progresses, the dynamics model can be gradually simplified to reduce computation and reach a longer planning horizon within the same computation budget. 
Different strategies to decrement the dynamics fidelity can be designed to efficiently balance the tradeoff between dynamics precision and computation overhead. We present our DDP implementation in Sec.~\ref{sec::implementations}. 


\section{Implementation}
\label{sec::implementations}

\begin{figure*}[!t]
  \centering  
  \includegraphics[width=\textwidth]{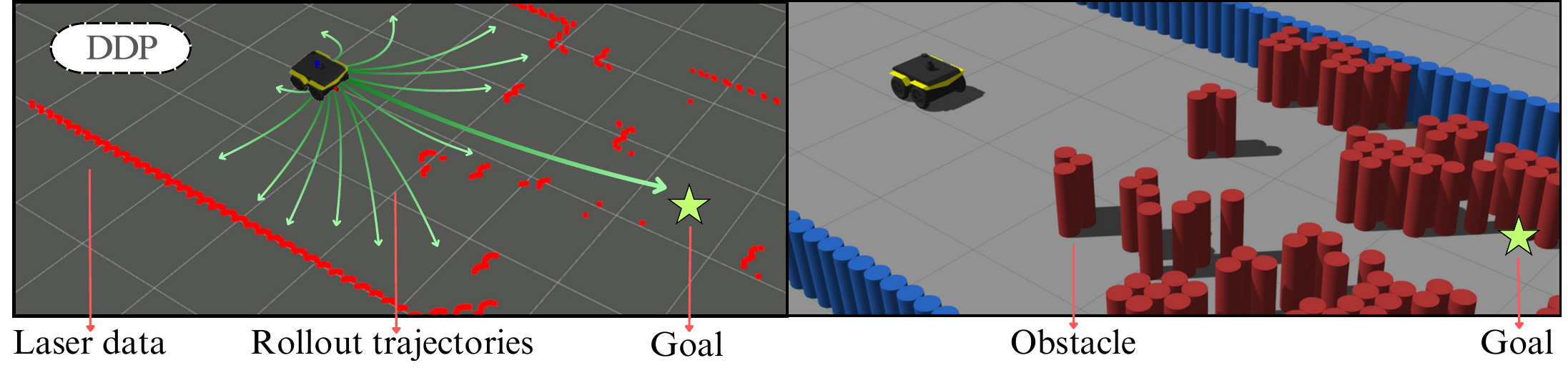}\\[-2mm]
  \caption{An example of our standalone DDP-based system navigating a BARN environment. Left: Visualization in RViz. Right: Visualization in Gazebo.}
  \label{fig:simulation}
\end{figure*}

Our implementation of DDP balances dynamics fidelity and onboard computation by adjusting both (1) the dynamics integration interval and (2) the number of robot state points for state-space collision checking at each time step. We augment our $\mathbb{SE}(2)$ robot state $(x_t, y_t, \psi_t)$ to $(x_t, y_t, \psi_t, x^1_t, y^1_t, \ldots, x^n_t, y^n_t)$, where $(x^i_t, y^i_t)_{i=1}^n$ are $n$ state points on the robot boundary for collision checking at step $t$. The robot dynamics parameters are $\theta_t=(\Delta_t, c^1_t, c^2_t, ..., c^n_t)$, where $\Delta_t$ denotes the dynamics integration interval at step $t$ and $\{c^i_t\}_{i=1}^n$ are binary indicators of whether the position of the $i$-th state point $(x^i_t, y^i_t)$ should be calculated and collision-checked at step $t$ ($c^i_t=1$) or not ($c^i_t=0$). 


To gradually increase the dynamics integration interval over the rollout time $\mathcal{T}$ (in seconds) we use the function: \begin{equation}
\textstyle
\Delta_t = \mathcal{T} \cdot \left[ \left(\frac{t+1}{T} \right)^p - \left(\frac{t}{T} \right)^p \right],
\label{eqn::ii}
\end{equation}
where $T$ is the number of time steps, and $p$ is a hyperparameter that controls the rate of change in the integration interval. At each integration interval, only the subset of the collision points defined by $\{c^i_t\}_{i=1}^n$ are checked. 
To gradually reduce the number of points checked along the trajectory ($N_t$) we use the function:  \begin{equation}
\textstyle
N_t = n \cdot \left( 1 - \left(\frac{t}{T}\right)^p \right).
\end{equation}
Only $\{c^i_t\}_{i=1}^{N_t}$ state points are calculated and collision-checked to decrement dynamics and save computation. 

Using this DDP design, we augment three different sampling-based motion planners with DDP: DWA~\cite{fox1997dynamic}, MPPI~\cite{williams2016aggressive}, and Log-MPPI~\cite{mohamed2022autonomous}. We also design a navigation system based solely on DDP.

\subsection{DDP-Augmented Planners}

\textbf{DWA} samples control sequences of linear and angular velocities based on the current state. It then generates trajectories using robot dynamics and evaluates them with a cost function using a set of collision points determined. In the original algorithm the dynamics and collision points are defined with a static $\theta$. In the DDP-augmented version $\theta$ is dynamic.


\textbf{MPPI} optimal control that leverages the path integral formulation to stochastic optimal control. It generates multiple trajectory samples using random perturbations of control inputs and evaluates their costs. Finally, the control input is computed as a weighted average based on these trajectories. When augmented by DDP, the trajectories are integrated with varying integration intervals, and different boundary points of the robot are utilized for collision checking,


\textbf{Log-MPPI} applies a logarithmic transformation to the cost function of MPPI, which combines stochastic sampling with model predictive control for real-time trajectory optimization. In Log-MPPI, the robot’s center point and local costmap are used for collision checking with a binary obstacle cost. DDP augmentation adapts the integration interval and costmap resolution to check collisions at varying boundary points.

All these planners and their DDP-augmented versions are run by themselves without any other complementary mechanisms, such as backup planners and recovery behaviors. 

\begin{figure}[t]
\centering
\includegraphics[width=0.98\columnwidth]{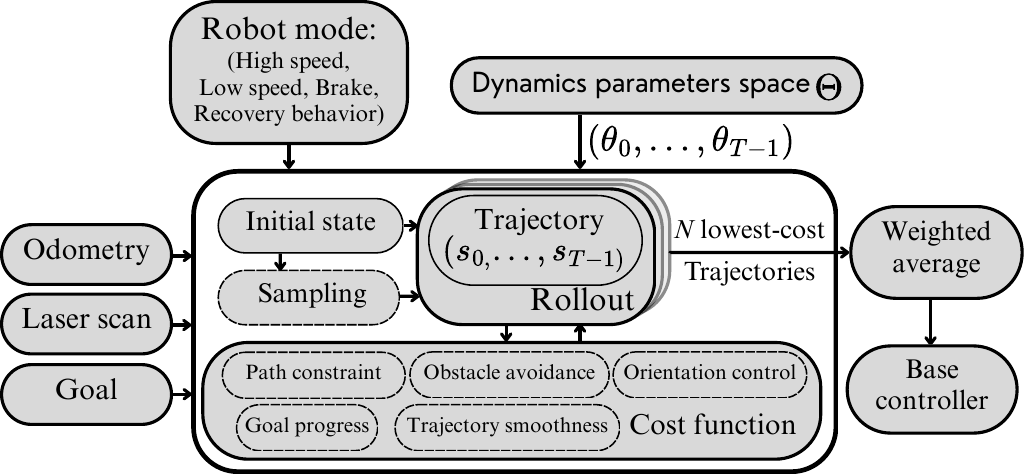}
\caption{Standalone DDP-based Navigation System.}
\label{fig:framework}
\end{figure}

\subsection{Standalone DDP-based Navigation System}
Our DDP-based system allows the robot to operate in high-speed, low-speed, braking, and recovery modes (including rotation and reverse). The robot starts in high-speed mode and shifts to low-speed when its linear velocity stays below a threshold for a set time, i.e., reducing the maximum linear and angular velocity limit for safe obstacle avoidance. If the linear velocity remains too low, it brakes before entering recovery behavior. Similarly, when the robot’s speed exceeds a specified threshold, it transitions from recovery to low-speed mode and then back to high-speed mode if necessary.

The navigation system's framework is shown in Fig.~\ref{fig:framework}. For each planning iteration, the robot first randomly samples linear and angular velocities with added noise. When the trajectory rollout begins, the system uses these velocity samples and the robot dynamics parameters to predict potential future trajectories and evaluate them against a comprehensive cost function. 
The cost function considers multiple factors: proximity to the goal, distance from obstacles, total path length, trajectory smoothness, and orientation relative to the goal. After all trajectories are evaluated, we retain only the $N=10$ collision-free trajectories with the lowest cost and use their linear and angular velocities to generate robot actions using a weighted average based on the trajectory cost.



\begin{table*}
    \centering
    \caption{Navigation Performance Comparison: Benchmarking the DDP-Based Navigation System and DDP-Augmented Planners across 300 Pre-Generated BARN Challenge Environments.}
    \begin{tabular}{lccc|ccc|ccc|ccc|ccc}
        \toprule
        \textbf{Method} & \multicolumn{3}{c}{\textbf{Task Success (\%)} $\uparrow$} & \multicolumn{3}{c}{\textbf{Avg. Time (s)} $\downarrow$} & \multicolumn{3}{c}{\textbf{Avg. Score} $\uparrow$} & \multicolumn{3}{c}{\textbf{Avg. Collision (\%)} $\downarrow$} & \multicolumn{3}{c}{\textbf{Avg. Timeout (\%)} $\downarrow$} \\
        \cmidrule(lr){2-4} \cmidrule(lr){5-7} \cmidrule(lr){8-10} \cmidrule(lr){11-13} \cmidrule(lr){14-16}
        & 1.0 & 1.5 & 2.0 & 1.0 & 1.5 & 2.0 & 1.0 & 1.5 & 2.0 & 1.0 & 1.5 & 2.0 & 1.0 & 1.5 & 2.0 \\
        \midrule 
        DWA & 90.75 & 81.83 & 62.73 & \textbf{10.00} & \textbf{07.14} & \textbf{05.89} & 0.454 & 0.409 & 0.313 & 9.25  & 18.17 & 37.27  & 0.00  &  0.00 & 0.00  \\
        DWA-DDP  & 92.42 & 83.61 & 68.56 & 10.13 & 07.57 & 06.66 & 0.462 & 0.418 & 0.343 & 7.58  & 16.39  & 31.44 & \textbf{0.00}  & \textbf{0.00}  & \textbf{0.00} \\
        \midrule
        MPPI & 86.06 & 82.72 & 71.01 & 10.98 & 07.64 & 06.10 & 0.430 & 0.414 & 0.355 & 13.82  & 17.28 & 28.99 & 0.12  & 0.00  & 0.00 \\
        MPPI-DDP & 89.13 & 86.18 & 79.26 & 11.33 & 08.37 & 07.12 & 0.446 & 0.431 & 0.396 & 10.76  & 13.71 & 20.74 & 0.11  & 0.11  & 0.00 \\
        \midrule
        Log-MPPI & 40.58 & 27.65 & 20.85 & 13.80 & 08.84 & 06.93 & 0.203 & 0.138 & 0.104 & 52.40  & 71.68 & 72.58 & 7.02  & 0.67  & 6.57 \\
        Log-MPPI-DDP & 50.61 & 44.59 & 40.47  & 12.72 & 08.39 & 06.49  & 0.253 & 0.223 & 0.202 & 43.37  & 54.52 & 54.29 & 6.02  & 0.89  & 5.24 \\
        \midrule
        DDP & \textbf{93.98} & \textbf{94.75} & \textbf{86.06} & 12.97 & 10.81 & 10.75 & \textbf{0.463} & \textbf{0.478} & \textbf{0.429} & \textbf{2.01}  & \textbf{2.90}  &  \textbf{1.67} & 4.01  & 2.35  & 12.27\\ 
        \bottomrule
        
    \end{tabular}
    \label{tab:DDP}
    
    \vspace{1pt}
    \parbox{\textwidth}{\raggedright Each experiment in each environment is performed three times; 1.0, 1.5, and 2.0 indicate the maximum linear velocity in m/s. Bold indicates best result. }

\end{table*}

\section{Experiments and Results}

\label{sec:ExpResults}
Experiments are conducted in both simulated and physical environments using a four-wheeled differential-drive Jackal robot equipped with a 2D Hokuyo LiDAR (270$^{\circ}$ field-of-view). We evaluate the three DDP-augmented planners and the standalone DDP-based navigation system, demonstrating that DDP significantly enhances planning performance.
\subsection{Experimental Setup}
\subsubsection{Methods for comparisons}
We implement DWA and MPPI in C++ and enhance them with DDP. Instead of using global or local costmaps in the ROS \texttt{move\_base} navigation stack, we construct the environment solely based on the current 2D laser scan. 
The cost function for these four planners considers progress toward the goal, the distance to the nearest obstacle along the trajectory, and the magnitude of angular velocity. In addition, our MPPI samples control inputs and evaluates all generated trajectories, selecting the best $N$ collision-free trajectories for weighted averaging. 
For the Log-MPPI planner, we use the public Python implementation\footnote{\url{https://github.com/IhabMohamed/log-MPPI_ros}} and augment it with DDP. These methods rely on costmaps to build the environment, and their cost function incorporates goal progress, a binary cost for collision avoidance, control effort, and trajectory smoothness.

\begin{figure}[t]
\centering
\includegraphics[width=0.95\columnwidth]{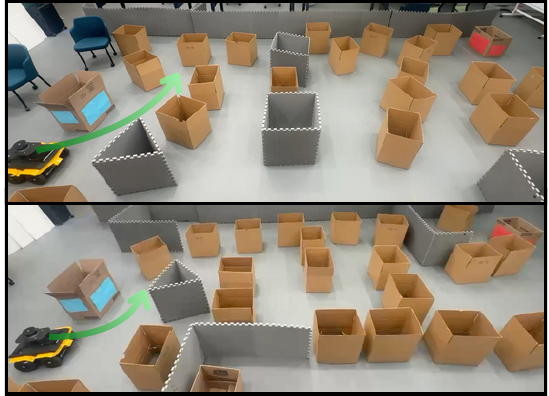}
\caption{Two Physical Test Environments.}
\label{fig:box}
\end{figure}

\subsubsection{Simulated environments}
The BARN dataset~\cite{perille2020benchmarking} comprises 300 simulated navigation environments generated by cellular automata. For the BARN challenge~\cite{xiao2022autonomous, xiao2023autonomous, xiao2024autonomous}, an additional set of 50 unpublished environments is introduced for testing. Fig.~\ref{fig:simulation} illustrates an example of our standalone DDP-based system navigating a BARN environment.

\subsubsection{Physical environment}
\begin{figure}[t]
\centering
\includegraphics[width=0.98\columnwidth]{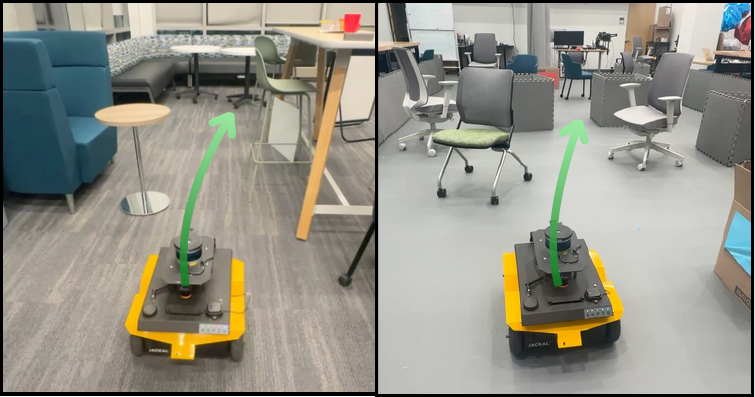}
\caption{Experiments in Real-World, Natural Cluttered Spaces.}
\label{fig:chair}
\end{figure}

We deploy in different physical environments to evaluate our DDP approaches, as shown in~Fig.~\ref{fig:box}. The blue box represents the starting point, while the red box indicates the goal. We calculate the robot's success rate, navigation progress, and running time. Fig.~\ref{fig:chair} shows another type of physical environment, where we test whether our DDP-based navigation system can navigate through obstacles in real-world, natural cluttered spaces.

\subsubsection{Computing Resources}
The experiments are conducted on an AMD Ryzen 9 5900X processor (3.7 GHz) running Ubuntu 20.04 with ROS Noetic onboard the robot. All code, except for the Log-MPPI implementation, is written in C++ and compiled using g++ 9.4.0.

\subsection{Simulation Result}
We compare DWA-DDP, MPPI-DDP, and Log-MPPI-DDP against their respective baseline methods to validate the effectiveness of DDP. Additionally, we compare our DDP-based system with those on the 2025 BARN Challenge leaderboard, where it ranks first in the simulation phase. The navigation score \( s_i \) for each environment \( i \) is computed as:
\begin{equation}
s_i = 1^{\text{success}} \times \frac{OT_i}{\text{clip}(AT_i, 2OT_i, 8OT_i)}
\nonumber
\end{equation}
where \( 1^{\text{success}} \) is an indicator variable set to 1 if the robot successfully reaches the navigation goal without collisions, and 0 otherwise. \( AT_i \) represents the actual traversal time, while \( OT_i \) denotes the optimal traversal time~\cite{xiao2022autonomous, xiao2023autonomous, xiao2024autonomous}. 

In Table~\ref{tab:DDP}, we compare the performance of DDP-based planners (DWA-DDP, MPPI-DDP, Log-MPPI-DDP, and DDP) against their baselines (DWA, MPPI, and Log-MPPI). The results demonstrate the effectiveness of our DDP paradigm, showing significant improvements in task success rates, navigation scores, and collision reduction across all speed settings (1.0 m/s, 1.5 m/s, and 2.0 m/s). While the performance gain varies among different planners and speed configurations, all DDP-augmented versions consistently outperform their baselines. A notable trade-off observed is the longer traversal time of the standalone DDP-based navigation system. This increased duration stems from its integrated safety mechanism, which dynamically reduces speed when navigating near obstacles—a critical safety feature absent in other planners. This deliberate design choice prioritizes collision avoidance and operational stability over raw speed. Regarding Log-MPPI's suboptimal performance, we identify two primary reasons: First, its reliance on costmap-based obstacle detection introduces perception errors, as nearby obstacles may be imprecise or prematurely cleared from its representation. Second, its cost function applies uniform weighting to all obstacles during trajectory averaging, resulting in poor planning outcomes when all sampled trajectories encounter collisions. Despite these inherent limitations, our DDP framework still enhances Log-MPPI's overall performance, further validating the robustness and adaptability of our approach across different planning architectures.

Table \ref{tab:hard} presents our second experiment, which evaluates the performance limits of these algorithms in the most challenging scenarios. We select the 25 most difficult BARN environments. The results further confirm the superior robustness of our DDP-augmented planners. Notably, Log-MPPI fails in all 25 environments and is therefore denoted as "/" in our results. To conclusively demonstrate the effectiveness of our standalone DDP-based system, we benchmark it against the top-performing algorithms on the 2025 BARN Challenge leaderboard, as shown in Table \ref{tab:competition}. Our DDP approach demonstrates strong performance, securing second place in both the simulation and real-world phases of the 2025 BARN Challenge.

\begin{table}[t]
    \centering
    \renewcommand{\arraystretch}{1.2} 
    \caption{Performance in the 25 Most Challenging BARN Challenge Environments.}
    \begin{tabular}{lcccc}
        \toprule
        \textbf{Method} & \textbf{Success (\%)} $\uparrow$ & \textbf{Avg. Time (s)} $\downarrow$ & \textbf{Avg. Score} $\uparrow$  \\
        \midrule
        DWA  & 36.67 & \textbf{7.60} & 0.183   \\
        DWA-DDP & 38.33 & 8.75 & 0.192 \\
        MPPI   & 26.67 & 8.96 & 0.133  \\
        MPPI-DDP  & 40.00 &  10.70 &  0.200  \\
        Log-MPPI   & / & / & / \\
        Log-MPPI-DDP  & 3.33 & 8.78 & 0.020  \\
        \midrule
        DDP  & \textbf{45.67}& 11.21 & \textbf{0.301}  \\
        \bottomrule
    \end{tabular}
    \label{tab:hard}
    
    \vspace{1pt}
    \parbox{\textwidth}{\raggedright The maximum linear velocity for all approaches is 1.5 m/s.}
    
\end{table}

\begin{table}[t]
    \centering
    \renewcommand{\arraystretch}{1.2} 
    \caption{Leaderboard Performance Comparison in the 2025 BARN Challenge in the 50 Unpublished Environments.}
    \begin{tabular}{lccc}
        \toprule
        \textbf{Method} & \textbf{Success (\%)} $\uparrow$ & \textbf{Avg. Time (s)} $\downarrow$ & \textbf{Avg. Score} $\uparrow$ \\
        \midrule
        INVENTEC  & 98.20 & 14.13 & 0.4206  \\
        KUL+FM  & \textbf{99.60} & 12.32 & 0.4641  \\
        AIMS & 96.00 & 9.70 & 0.4723 \\
        LiCS-KI~\cite{damanik2024lics}   & 95.40 & \textbf{7.55} & 0.4762 \\
        DDP   & 99.00 & 7.67 & 0.4873 \\
        FSMT   & 98.20 & 8.65 & \textbf{0.4878} \\
        \bottomrule
    \end{tabular}
    
    \label{tab:competition}
\end{table}

\begin{table}[t]
    \centering
    \renewcommand{\arraystretch}{1.2} 
    \caption{Performance in Two Real-World Environments.}
    \begin{tabular}{lcccc}
        \toprule
        \textbf{Method} & \textbf{Success} $\uparrow$ & \textbf{Avg. Progress} (\%) $\uparrow$ & \textbf{Avg. Time (s)} $\downarrow$  \\
        \midrule
        DWA  & 0/10 & 70.52 $\pm$ 2.17 & /   \\ 
        DWA-DDP & 0/10 & 77.13 $\pm$ 3.25 & / \\
        MPPI   & 0/10 & 76.25 $\pm$ 2.54 & /  \\
        MPPI-DDP  & 2/10 &  80.70 $\pm$ 4.52 & \textbf{27.05 $\pm$ 2.12} \\
        \midrule
        DDP  & \textbf{10/10} & \textbf{100.00 $\pm$ 0.00} & 28.04 $\pm$ 1.23 \\
        \bottomrule
    \end{tabular}
    \label{tab:real_world}
    
    \vspace{4pt}
    \parbox{\textwidth}{\raggedright The maximum linear velocity for all approaches is 1.5 m/s.}
    
\end{table}

\begin{table}[t]
    \centering
    \renewcommand{\arraystretch}{1.2} 
    \caption{Parameter Sensitivity Study: Impact of Integration Interval on DDP Performance.}
    \begin{tabular}{lcccc}
        \toprule
        \textbf{Method} & \textbf{Success (\%)} $\uparrow$ & \textbf{Avg. Time (s)} $\downarrow$ & \textbf{Avg. Score} $\uparrow$  \\
        \midrule
        DDP-P = 1.2 & 98.30 & 7.52 & 0.481   \\
        DDP-P = 1.4 & 98.32 & \textbf{7.37} & \textbf{0.484}   \\
        DDP-P = 1.7  & 97.31 & 9.24 & 0.478  \\
        DDP-P = 2.0  & \textbf{98.65} &  8.11 &  0.481  \\
        \bottomrule
    \end{tabular}
    
    \label{tab:ab}
    \vspace{4pt}
    \parbox{\textwidth}{\raggedright The maximum linear velocity for all approaches is 1.5 m/s.}
    
\end{table}

\subsection{Physical Environment}
For real-world experiments (Fig.\ref{fig:box}), we design two types of complex environments and test five navigation methods: DWA, DWA-DDP, MPPI, MPPI-DDP, and DDP. Due to the poor performance of Log-MPPI in previous experiments and safety considerations, we exclude it from physical testing. Each method is evaluated five times in each environment. For successful task completions, we calculate the average time; for failed attempts, we measure the average navigation progress. As shown in Table \ref{tab:real_world}, our DDP-augmented methods demonstrate superior ability to make significant progress even in failed attempts. For the scenarios in Fig.~\ref{fig:chair}, we conduct experiments exclusively with our DDP method, which successfully navigates through both natural environments.

\subsection{Parameter Sensitivity Study} Table~\ref{tab:ab} presents an parameter sensitivity study on our standalone DDP system, focusing on the $p$ value to determine the integration interval (Eqn.~\ref{eqn::ii}). We find that adjusting the integration interval has minimal impact on the experimental results, as all variations still achieve strong performance.

\section{Conclusion}
\label{sec:Conclusions}

In this work, we introduce DDP, a novel paradigm that integrates dynamic constraints into the entire planning process, overcoming the limitations of traditional decomposed navigation frameworks. Unlike conventional approaches that either fully incorporate or completely ignore robot dynamics at different planning levels, DDP starts with high-fidelity dynamics modeling and gradually reduces its complexity to balance computational efficiency and dynamic feasibility. We validate the effectiveness of DDP by augmenting multiple planners, including DWA, MPPI, and Log-MPPI, demonstrating improved success rates and efficiency in both simulated and real-world experiments. 
Additionally, we develop a standalone DDP-based planner, which secures first place in the 2025 BARN Challenge simulation phase, achieving superior performance in the most challenging navigation scenarios. The consistent improvements observed across all tested planners highlight DDP's potential as a general framework for enhancing robot navigation systems. Future work will focus on extending DDP to more challenging real-world scenarios, incorporating adaptive fidelity reduction strategies, and exploring its integration with learning-based methods to further optimize motion planning for autonomous robots. 
\section*{acknowledgements}
This work has taken place in the RobotiXX Laboratory at George Mason University. RobotiXX research is supported by National Science Foundation (NSF, 2350352), Army Research Office (ARO, W911NF2320004, W911NF2420027, W911NF2520011), Air Force Research Laboratory (AFRL), US Air Forces Central (AFCENT), Google DeepMind (GDM), Clearpath Robotics, Raytheon Technologies (RTX), Tangenta, Mason Innovation Exchange (MIX), and Walmart.

\IEEEpeerreviewmaketitle

\bibliographystyle{IEEEtran}
\bibliography{ref.bib}

\end{document}